\documentclass{article}
\usepackage[dvips]{graphicx}
\usepackage{nips07submit_e,times}

\usepackage{amssymb,amsfonts,amsmath,amsthm,amscd,mathrsfs,dsfont,bbm}

\newtheorem{propo}{Proposition}[section]
\newtheorem{lemma}[propo]{Lemma}

\newtheorem{thm}[propo]{Theorem}

\def\cE{{\cal E}}
\def\E{{\mathbb E}}
\def\uX{\underline{X}}
\def\eC{\widehat{C}}
\def\ux{\underline{x}}
\def\alg{{\sf Alg}}
\def\thres{{\sf Thr}}
\def\ind{{\sf Ind}}
\def\indD{{\sf IndD}}
\def\rlr{{\sf Rlr}}
\def\prob{{\mathbb P}}
\def\hprob{\widehat{\mathbb P}}
\def\atanh{{\rm atanh}}
\def\dr{{\partial r}}
\def\di{{\partial i}}
\def\score{\mbox{\sc Score}}
\def\utheta{\underline{\theta}}
\def\hutheta{\hat{\underline{\theta}}}

\def\hz{\hat{z}}
\def\ua{\underline{a}}
\def\eps{\epsilon}
\def\psucc{{\rm P}_{\rm succ}}

\def\Tree{{\sf T}}
\def\dtree{\partial{\sf T}}
\def\Ball{{\sf B}}
\def\dist{{\rm dist}}

\newcommand{\reals}{{\mathds R }}

\newcommand{\naturals}{{ \mathds N}}

\title{Which graphical models are difficult to learn?}

\author{
Jos\'{e} Bento \\
Department of Electrical Engineering\\
Stanford University\\
\texttt{jbento@stanford.edu} \\
\And
Andrea Montanari\\
Department of Electrical Engineering and\\
Department of Statistics\\
Stanford University\\
\texttt{montanari@stanford.edu} \\
}

\begin{document}

\maketitle

\begin{abstract}
We consider the problem of learning the structure of
Ising models (pairwise binary Markov random fields) from i.i.d. samples.
While several methods have been proposed to accomplish this task,
their relative merits and limitations remain somewhat obscure.
By analyzing a number of concrete examples, we show that low-complexity 
algorithms systematically fail when the Markov random field 
develops long-range correlations. More precisely, this phenomenon 
appears to be related to the Ising model phase transition 
(although it does not coincide with it).
\end{abstract}

\section{Introduction and main results}

Given a graph $G= (V=[p],E)$, and a positive parameter $\theta>0$
the \emph{ferromagnetic Ising model on $G$} is the pairwise
Markov random field 
\begin{eqnarray}
\mu_{G,\theta}(\ux) =\frac{1}{Z_{G,\theta}}\,
\prod_{(i,j)\in E} e^{\theta x_i x_j}\label{eq:IsingModel}
\end{eqnarray}
over binary variables $\ux = (x_1,x_2,\dots,x_p)$. Apart from being 
one of the most studied models in statistical mechanics, the Ising model
is a prototypical undirected graphical model, with applications in 
computer vision, clustering and spatial statistics. Its obvious generalization
to edge-dependent parameters $\theta_{ij}$, $(i,j)\in E$ is of interest 
as well, and will be introduced in Section \ref{sec:Pseudo}.
(Let us stress that we follow the statistical mechanics convention
of calling (\ref{eq:IsingModel}) an Ising model for any graph $G$.)

In this paper we study the following structural learning problem:
\emph{Given $n$ i.i.d. samples $\ux^{(1)}$, $\ux^{(2)}$,\dots, $\ux^{(n)}$
with distribution $\mu_{G,\theta}(\,\cdot\,)$, reconstruct the 
graph $G$.} For the sake of simplicity, we assume that the parameter 
$\theta$ is known, and that $G$ has no double edges (it is a 
`simple' graph).

The graph learning problem is solvable with unbounded
sample complexity, and computational resources \cite{abbeel}. The question we
address is: for which classes of graphs and values of the parameter
$\theta$ is the problem solvable under appropriate complexity
constraints? More precisely, given an algorithm $\alg$, a graph
$G$, a value $\theta$ of the model parameter, and a small 
$\delta>0$, the sample complexity is defined as
\begin{eqnarray}
n_{\alg}(G,\theta) \equiv \inf\left\{ n\in\naturals:\, 
\prob_{n,G,\theta}\{\alg(\ux^{(1)},\dots,\ux^{(n)})=G\}\ge 1-\delta\right\}\, ,
\end{eqnarray}
where $\prob_{n,G,\theta}$ denotes probability with respect to $n$ 
i.i.d. samples with distribution $\mu_{G,\theta}$. 
Further, we let $\chi_{\alg}(G,\theta)$ denote the number of operations
of the algorithm $\alg$, when run on $n_{\alg}(G,\theta)$ 
samples.\footnote{For the algorithms analyzed in this paper, 
the behavior of $n_{\alg}$ and $\chi_{\alg}$ does not change significantly if 
we require only `approximate' reconstruction (e.g. in graph distance).}

The general problem is therefore to characterize the functions
$n_{\alg}(G,\theta)$ and $\chi_{\alg}(G,\theta)$, in particular for 
an optimal choice of the algorithm. General bounds on $n_{\alg}(G,\theta)$ 
have been given in \cite{martin_info_limits,santhanam_info_limits},
under the assumption of unbounded computational resources. 
A general charactrization of how well low complexity algorithms can perform
is therefore lacking. Although we cannot prove such a general characterization, 
in this paper we estimate $n_{\alg}$ and $\chi_{\alg}$ 
for a number of graph models, as a function of $\theta$, and unveil a 
fascinating universal pattern: 
\emph {when the model (\ref{eq:IsingModel}) develops long range 
correlations,  low-complexity algorithms fail.}
Under the Ising model, the variables $\{x_i\}_{i\in V}$ 
become strongly correlated for $\theta$ large. For a large class of graphs with degree bounded by $\Delta$, 
this phenomenon corresponds to a phase transition beyond some
critical value of $\theta$ uniformly bounded in $p$,
with typically $\theta_{\rm crit}\le {\rm const.}/\Delta$. In the examples 
discussed below, the failure of low-complexity algorithms appears to be related to this phase transition (although it does not coincide with it).
%
%
\subsection{A toy example: the thresholding algorithm}

In order to illustrate the interplay between graph structure,
sample complexity and interaction strength $\theta$, it is
instructive to consider a warmup example. The thresholding 
algorithm reconstructs $G$ by thresholding the empirical correlations
\begin{eqnarray}
\eC_{ij} \equiv \frac{1}{n}\sum_{\ell =1}^{n}x^{(\ell)}_ix^{(\ell)}_j\, 
\;\;\;\;\;\;\;\mbox{for $i,j\in V$}. 
\end{eqnarray}
\begin{tabular}{ll}
\hline
\multicolumn{2}{l}{ {\sc Thresholding}( samples $\{x^{(\ell)}\}$, threshold 
$\tau$ )}\\
\hline
1: & Compute the empirical correlations $\{\eC_{ij}\}_{(i,j)\in V\times V}$;\\
2: & For each $(i,j)\in V\times V$\\
3: & \phantom{aaa}If $\eC_{ij}\ge \tau$, set $(i,j)\in E$;\\
\hline
\end{tabular}

\phantom{a}

We will denote this algorithm by $\thres(\tau)$. Notice that its complexity
is dominated by the computation of the empirical correlations,
i.e. $\chi_{\thres(\tau)} = O(p^2n)$. 
The sample complexity $n_{\thres (\tau)}$ can be bounded for specific
classes of graphs as follows
(the proofs are straightforward and omitted from this paper).
\begin{thm}
If $G$ has maximum degree $\Delta > 1$ and if $\theta < 
\atanh (1/(2\Delta))$ then there exists $\tau = \tau(\theta)$ such that
\begin{eqnarray}
n_{\thres (\tau)}(G,\theta) \le \frac{8}{(\tanh \theta - \frac{1}{2\Delta}  )^{2}} 
\;\log \frac{2p}{\delta}\,.
\end{eqnarray}
Further, the choice $\tau(\theta) = (\tanh \theta + (1/2\Delta))/2$ achieves
this bound.\label{th:tresh2}
\end{thm}
\begin{thm}
There exists a numerical constant $K$ such that the following is true.
If $\Delta > 3$ and $\theta> K/\Delta$, there are graphs of bounded degree $\Delta$ such that for any $\tau$, $n_{\thres(\tau)} = \infty$, i.e. the thresholding algorithm always fails with high probability. 
\label{th:tresh3}
\end{thm}
These results confirm the idea that the failure of low-complexity algorithms 
is related to long-range correlations in the underlying graphical model.
If the graph $G$ is a tree, then correlations between far apart variables
$x_i$, $x_j$ decay exponentially with the distance between vertices
$i$, $j$. The same happens on bounded-degree graphs if 
$\theta\le {\rm const.}/\Delta$. However, for $\theta > {\rm const.}/\Delta$,
there exists families of bounded degree graphs with long-range correlations.

%
%
\subsection{More sophisticated algorithms}

In this section we characterize $\chi_{\alg}(G,\theta)$ and 
$n_{\alg}(G,\theta)$ for more advanced algorithms. We again obtain
very distinct behaviors of these algorithms depending on 
long range correlations. Due to space limitations, 
we focus on two type of algorithms and only outline the proof
of our most challenging result, namely Theorem \ref{th:mart2}.

In the following we denote by $\di$ the neighborhood of a node 
$i\in G$ ($i \notin \partial i$), and assume the degree to be bounded: $|\di|\le \Delta$. 

\subsubsection{Local Independence Test}

A recurring approach to structural learning consists in exploiting
the conditional independence structure encoded by the graph
\cite{abbeel,mossel,Csiszar,Friedman}. 

Let us consider, to be definite,
the approach of \cite{mossel}, specializing it to the model 
(\ref{eq:IsingModel}). Fix a vertex $r$, whose neighborhood 
we want to reconstruct,  and consider the conditional
distribution of $x_r$ given its 
neighbors\footnote{If $\ua$ is a vector and $R$ is a set of indices then  
we denote by $\ua_R$ the vector formed by the components of $\ua$ with index 
in $R$.}:
$\mu_{G,\theta}(x_r|\ux_{\dr})$. Any change of $x_i$, $i\in\dr$,
produces a change in this distribution which is bounded away from $0$.
Let $U$ be a candidate neighborhood, and assume
 $U\subseteq\dr$. Then changing the value of $x_j$, $j \in U$ 
 will produce a noticeable change in the marginal of $X_r$, even
if we condition   on the remaining values in $U$ and in any $W$,
$|W|\le \Delta$. 
On the other hand, if $U\nsubseteq \dr$, then it is  possible to find $W$ 
(with $|W|\le \Delta$) and a node $i \in U$ such that, changing its value 
after fixing all other values in $U \cup W$ will produce no noticeable 
change in the conditional marginal. (Just choose $i \in U \backslash \dr$ 
and $W = \dr\backslash U$). This procedure allows us to distinguish subsets 
of $\dr$ from other sets of vertices, thus motivating the following algorithm.

\phantom{a}

\begin{tabular}{ll}
\hline
\multicolumn{2}{l}{ {\sc Local Independence Test}( samples $\{x^{(\ell)}\}$, thresholds $(\epsilon,\gamma)$ )}\\
\hline
1: & Select a node $r \in V$;\\
2: & Set as its neighborhood the largest candidate neighbor $U$ of\\
   & size at most $\Delta$ for which the score function $\score(U) > \epsilon/2$;\\
3: & Repeat for all nodes $r \in V$;\\
\hline
\end{tabular}

\phantom{a}

The score function $\score(\,\cdot\,)$ depends on $(\{x^{(\ell)}\},\Delta,\gamma)$ and is defined as follows,
\begin{align}
\min_{W,j} \max_{x_i,\underline{x}_W,\underline{x}_U,x_j} &|\hprob_{n,G,\theta}\{X_i = x_i | \underline{X}_W = \underline{x}_W,\underline{X}_{U} = \underline{x}_{U}\} - \nonumber \\
&\hprob_{n,G,\theta}\{X_i = x_i | \underline{X}_W = \underline{x}_W,\underline{X}_{U \backslash j} = \underline{x}_{U \backslash j}, X_j = x_j\}|\, .
\label{eq:ScoreDef}
\end{align}
In the minimum, $|W| \leq \Delta$ and $j \in U$. In the maximum, the values must be such that
\begin{align}
\hprob_{n,G,\theta}\{\underline{X}_W = \underline{x}_W,\underline{X}_{U} = \underline{x}_{U}\} > \gamma / 2, \nonumber \;\;\;\;\;
\hprob_{n,G,\theta}\{\underline{X}_W = \underline{x}_W,\underline{X}_{U \backslash j} = \underline{x}_{U \backslash j}, X_j = x_j\} > \gamma / 2
\end{align}
$\hprob_{n,G,\theta}$ is the empirical distribution calculated from the samples $\{x^{(\ell)}\}$. We denote this algorithm by $\ind(\epsilon,\gamma)$. The search over candidate neighbors $U$, the search for minima and maxima in the  computation of the $\score(U)$ and the computation of $\hprob_{n,G,\theta}$ all contribute for $\chi_{\ind}(G,\theta)$.
 
Both theorems that follow are consequences of the analysis of \cite{mossel}.
\begin{thm}
Let $G$ be a graph of bounded degree $\Delta\ge 1$. For every $\theta$ there exists $(\epsilon, \gamma)$, and a numerical constant $K$, such that 
\begin{align*}
n_{\ind (\epsilon,\gamma)}(G,\theta) \le  \frac{100 \Delta }{\epsilon^2 \gamma^4} \log \frac{2p}{\delta}\, ,\;\;\;\;\;
\chi_{\ind_{(\epsilon,\gamma)}}(G,\theta) \le K\, (2p)^{2\Delta +1} \log p\, .
\end{align*}
More specifically, one can take 
$\epsilon = \frac{1}{4} \sinh (2\theta)$, $\gamma = e^{-4 \Delta \theta} \; 2^{-2\Delta}$.\label{th:mossel1}
\end{thm}
This first result  implies in particular that $G$ can be reconstructed with 
polynomial complexity for any bounded $\Delta$. However, the degree of
such polynomial is pretty high and non-uniform in $\Delta$. This makes the
above approach impractical.

A way out was proposed in \cite{mossel}. The idea is to identify 
a set of `potential neighbors' of vertex $r$ via thresholding:
\begin{equation}
B(r) = \{i \in V: \eC_{r i} > \kappa / 2\}\, ,
\end{equation}
For each node $r \in V$, we evaluate $\score(U)$ by restricting the minimum in
Eq.~(\ref{eq:ScoreDef}) over $W\subseteq B(r)$, and search only over
$U\subseteq B(r)$. We call this algorithm $\indD(\epsilon,\gamma,\kappa)$.
The basic intuition here is that $C_{ri}$ decreases rapidly 
with the graph distance between vertices $r$ and $i$. As mentioned
above, this is true at small $\theta$.
\begin{thm}
Let $G$ be a graph of bounded degree $\Delta\ge 1$. Assume that 
$\theta < K / \Delta$ for some small enough constant $K$. 
Then there exists $\epsilon, \gamma,\kappa$ such that 
\begin{align*}
n_{\indD (\epsilon,\gamma,\kappa)}(G,\theta) \le  8(\kappa^2+8^{\Delta}) 
\log \frac{4p}{\delta} \, ,\;\;\;\;\;
\chi_{\indD_{(\epsilon,\gamma,\kappa)}}(G,\theta) \le K' p \Delta^{\Delta \frac{\log(4/ \kappa)}{\alpha}} + K' \Delta p^2 \log p\, .
\end{align*}
More specifically, we can take 
$\kappa = \tanh \theta$, $\epsilon = \frac{1}{4} \sinh (2\theta)$
and $\gamma = e^{-4 \Delta \theta} \; 2^{-2\Delta}$. \label{th:mossel2}
\end{thm}

%
%
\subsubsection{Regularized Pseudo-Likelihoods}
\label{sec:Pseudo}

A different approach to the learning problem 
consists in maximizing an appropriate empirical likelihood function \cite{martin,martin2,Tibshirani,Ghaoui,tibshirani_lasso}. 
To control the fluctuations caused by the limited number of samples,
and select sparse graphs a regularization term is often added \cite{martin,martin2,Tibshirani,Ghaoui,Yuan,Meinshausen,tibshirani_lasso}.

As a specific low complexity implementation of this idea,
we consider the $\ell_1$-regularized pseudo-likelihood method
of \cite{martin}. For each node $r$, the following likelihood 
function is considered 
\begin{equation}
L (\utheta;\{x^{(\ell)}\}) = - \displaystyle{ \frac{1}{n} \sum_{\ell = 1}^n {\log  \prob_{n,G,\utheta} (x_r^{(\ell)}| x_{\backslash r}^{(\ell)}) }}
\end{equation}
where $\ux_{\backslash r}=\ux_{V\setminus r} 
= \{x_i:\, i\in V\setminus r\}$ 
is the vector of all variables except  $x_r$ and 
$\prob_{n,G,\utheta}$ is defined from the following extension of (\ref{eq:IsingModel}),
\begin{eqnarray}
\mu_{G,\utheta}(\ux) =\frac{1}{Z_{G,\utheta}}\,
\prod_{i,j \in V} e^{\theta_{ij} x_i x_j}\label{eq:IsingModel2}
\end{eqnarray}
where $\utheta = \{\theta_{ij}\}_{i,j\in V}$ is a vector of real 
parameters. Model (\ref{eq:IsingModel}) corresponds to 
$\theta_{ij} = 0, \; \forall (i,j) \notin E$ and $\theta_{ij} = \theta, \; \forall (i,j) \in E$.

The function $L (\utheta;\{x^{(\ell)}\})$ depends only on 
$\utheta_{r,\cdot} = \{\theta_{rj}, \, j \in \dr \}$ and is used to estimate 
the neighborhood of each node by the following algorithm, $\rlr(\lambda)$,

\phantom{a}

\begin{tabular}{ll}
\hline
\multicolumn{2}{l}{ {\sc Regularized Logistic Regression}( samples $\{x^{(\ell)}\}$, regularization $(\lambda)$)}\\
\hline
1: & Select a node $r \in V$;\\
2: & Calculate $\displaystyle \hat{\utheta}_{r, \cdot} = \arg \min_{\utheta_{r,\cdot} \in \reals^{p-1} } \{ L(\utheta_{r,\cdot};\{x^{(\ell)}\}) + \lambda || \utheta_{r,\cdot} ||_1 \}$;\\
3: & \phantom{aaa} If $\hat{\theta}_{rj} >0$, set $(r,j) \in E$;\\

\hline
\end{tabular}

\phantom{a}

Our first result shows that $\rlr(\lambda)$ indeed reconstructs 
$G$ if $\theta$ is sufficiently small.
\begin{thm}
There exists numerical constants $K_1$, $K_2$, $K_3$,
such that the following is true.
Let $G$ be a graph with degree bounded by $\Delta\ge 3$. 
If $\theta \le K_1/\Delta$, then there exist $\lambda$ such that
\begin{equation}
n_{\rlr (\lambda)}(G,\theta) \leq K_2 \, \theta^{-2} \, \Delta\,  
\log  \frac{8p^2}{\delta} \, .
\end{equation}
Further, the above holds with $\lambda = K_3 \, \theta \, \Delta^{-1/2}$.\label{th:mart1}
\end{thm}
This theorem is proved by noting that for $\theta \le K_1/\Delta$ correlations decay exponentially, which makes all conditions  in Theorem 1 of \cite{martin} (denoted there by A1 and A2) hold, and then computing the probability of success as a function of $n$, while strenghtening the error bounds of \cite{martin}.

In order to prove a converse to the above result, we need
to make some assumptions on  $\lambda$.
Given $\theta>0$, we say that $\lambda$ is `reasonable
for that value of $\theta$ if the following conditions old:
$(i)$ $\rlr(\lambda)$ is successful with probability larger than $1/2$
on any star graph  (a graph composed by a vertex $r$ connected to 
$\Delta$ neighbors, plus isolated vertices); $(ii)$
$\lambda\le \delta(n)$ for some sequence $\delta(n)\downarrow 0$.
\begin{thm}
There exists a numerical constant $K$ such that the following happens.
If $\Delta>3$, $\theta> K/\Delta$, then there exists graphs $G$ 
of degree bounded by $\Delta$ such that for all reasonable 
$\lambda$, $n_{\rlr(\lambda)}(G) = \infty$, i.e.
regularized logistic regression fails with high probability.
\label{th:mart2}
\end{thm}
The graphs for which regularized logistic regression fails
are not contrived examples. Indeed we will prove that the
claim in the last theorem holds with high probability when $G$
is a uniformly random graph of regular degree $\Delta$.

The proof Theorem \ref{th:mart2} is based on showing that an appropriate 
\emph{incoherence condition} is necessary for $\rlr$ to successfully 
reconstruct $G$. The analogous result was proven in \cite{zhao} for model 
selection using the Lasso. In this paper we show that 
such a  condition is also necessary when the underlying model is an 
Ising model. 
Notice that, given the graph $G$, checking the incoherence condition
is NP-hard for general (non-ferromagnetic) Ising model, and 
requires significant computational effort even in the ferromagnetic case. 
Hence the incoherence condition does not provide, by itself,
a clear picture of which graph structure are difficult to learn.
We will instead show how to evaluate it on specific graph families.

Under the restriction  $\lambda \rightarrow 0$ the solutions given by 
$\rlr$ converge to $\utheta^*$ with $n$ \cite{martin}. Thus, for large $n$ 
we can expand $L$ around $\utheta^*$ to second order in $(\utheta-\utheta^*)$. 
When we add the regularization term to $L$ we obtain a quadratic model 
analogous the Lasso plus the error term due to the quadratic approximation. 
It is thus not surprising that, when $\lambda \rightarrow 0$ the 
incoherence condition introduced for the Lasso  in \cite{zhao}
is also relevant for the Ising model.

%
%
\section{Numerical experiments}

In order to explore the practical relevance of the above results,
we carried out extensive numerical simulations using the 
regularized logistic regression algorithm $\rlr(\lambda)$.
Among other learning algorithms, $\rlr(\lambda)$
strikes a good balance of complexity and performance.
Samples from the Ising model (\ref{eq:IsingModel}) where generated 
using Gibbs sampling (a.k.a. Glauber dynamics). Mixing time can be 
very large for $\theta\ge \theta_{\rm crit}$, and was estimated using
the time required for the overall bias to change sign (this is a quite 
conservative estimate at low temperature). Generating the samples
$\{\ux^{(\ell)}\}$ was indeed the bulk of our computational effort and
took about $50$ days CPU time on Pentium Dual Core processors 
(we show here only part of these data). 
Notice that $\rlr(\lambda)$ had been tested in \cite{martin}
only on tree graphs $G$, or in the weakly coupled
regime $\theta<\theta_{\rm crit}$. 
In these cases sampling from the Ising model is easy,
but structural learning is also intrinsically easier.

\begin{figure}
\phantom{a}\hspace{-1.2cm}\includegraphics[width=0.61\linewidth]{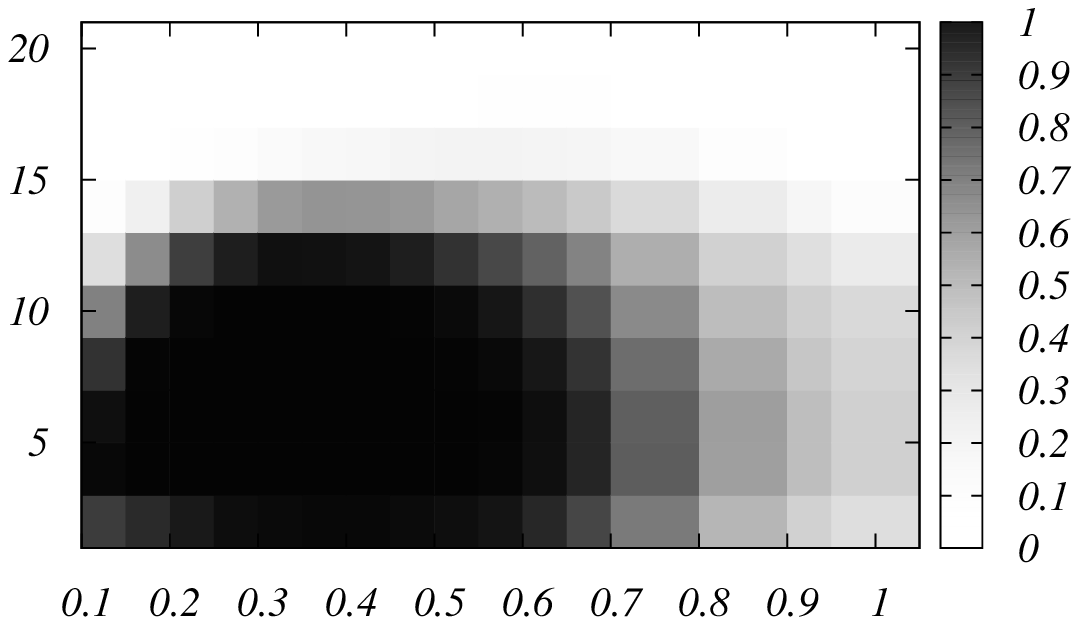}
\hspace{-0.1cm}\includegraphics[width=0.52\linewidth]{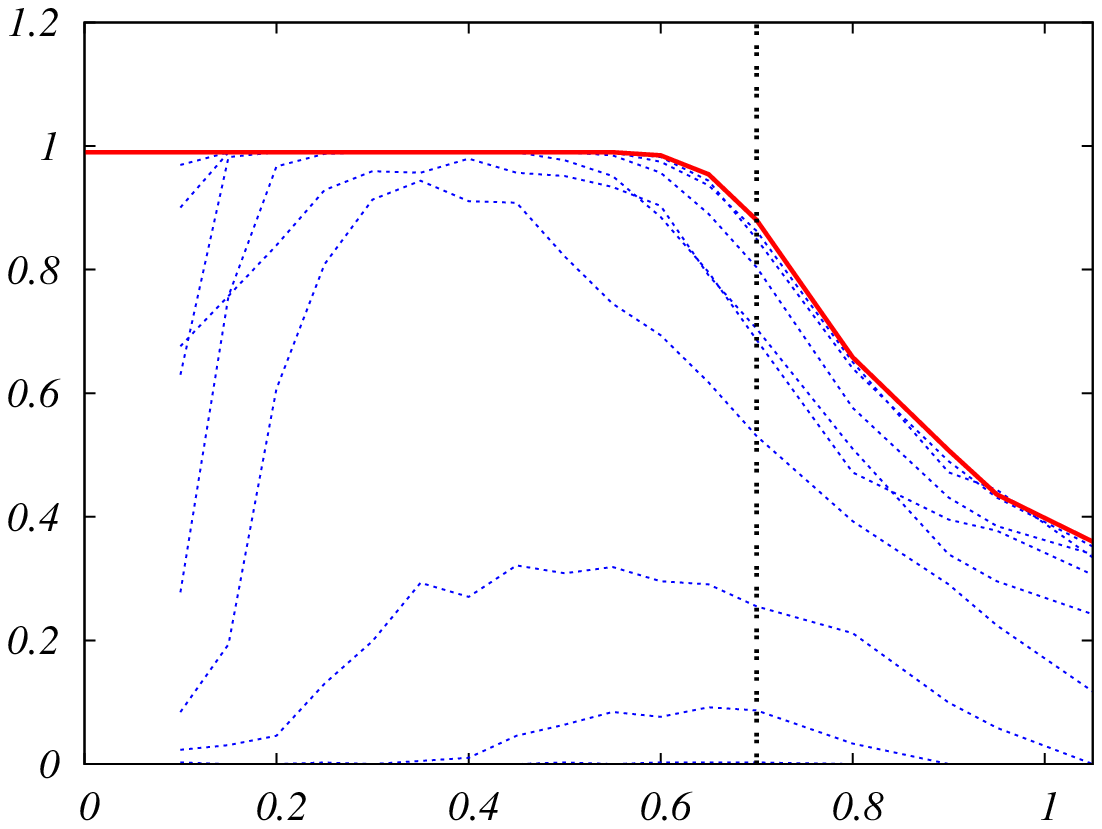}
\put(-435,80){$\lambda_0$}
\put(-330,10){$\theta$}
\put(-75,5){$\theta_{crit}$}
\put(-100,-5){$\theta$}
\put(-210,65){P${}_{\rm succ}$}
\caption{Learning random subgraphs of a $7\times 7$ ($p=49$) 
two-dimensional grid from $n=4500$ Ising models samples, using regularized 
logistic regression. Left: success probability as a function of
the model parameter $\theta$ and of the regularization parameter $\lambda_0$
(darker corresponds to highest probability). Right: the same data 
plotted for several choices of $\lambda$ versus $\theta$. The vertical line
corresponds to the model critical temperature. The thick line is
an envelope of the curves obtained for different $\lambda$, and 
should correspond to optimal regularization.}\label{fig:Grid}
\end{figure}

Figure reports the success probability of $\rlr(\lambda)$
when applied to random subgraphs of a $7\times 7$
two-dimensional grid. Each such graphs was obtained by removing each edge
independently with probability $\rho = 0.3$. Success probability was
estimated by applying $\rlr(\lambda)$ to each vertex of $8$ graphs 
(thus averaging over $392$ runs of $\rlr(\lambda)$), using $n=4500$ samples.
We scaled the regularization  parameter as 
$\lambda = 2\lambda_0 \theta (\log p/n)^{1/2}$ (this choice is motivated by 
the algorithm analysis and is empirically the most satisfactory),
and searched over $\lambda_0$.

The data clearly illustrate the phenomenon discussed.
Despite the large number of samples $n\gg \log p$,
when $\theta$ crosses a threshold, the algorithm
starts performing poorly irrespective of $\lambda$.
Intriguingly, this threshold is not far from the critical point of
the Ising model on a randomly diluted grid $\theta_{\rm crit}(\rho=0.3)
\approx 0.7$ \cite{Zobin,fisher}.

\begin{figure}
\phantom{a}\hspace{-0.5cm}\includegraphics[width=0.52\linewidth]{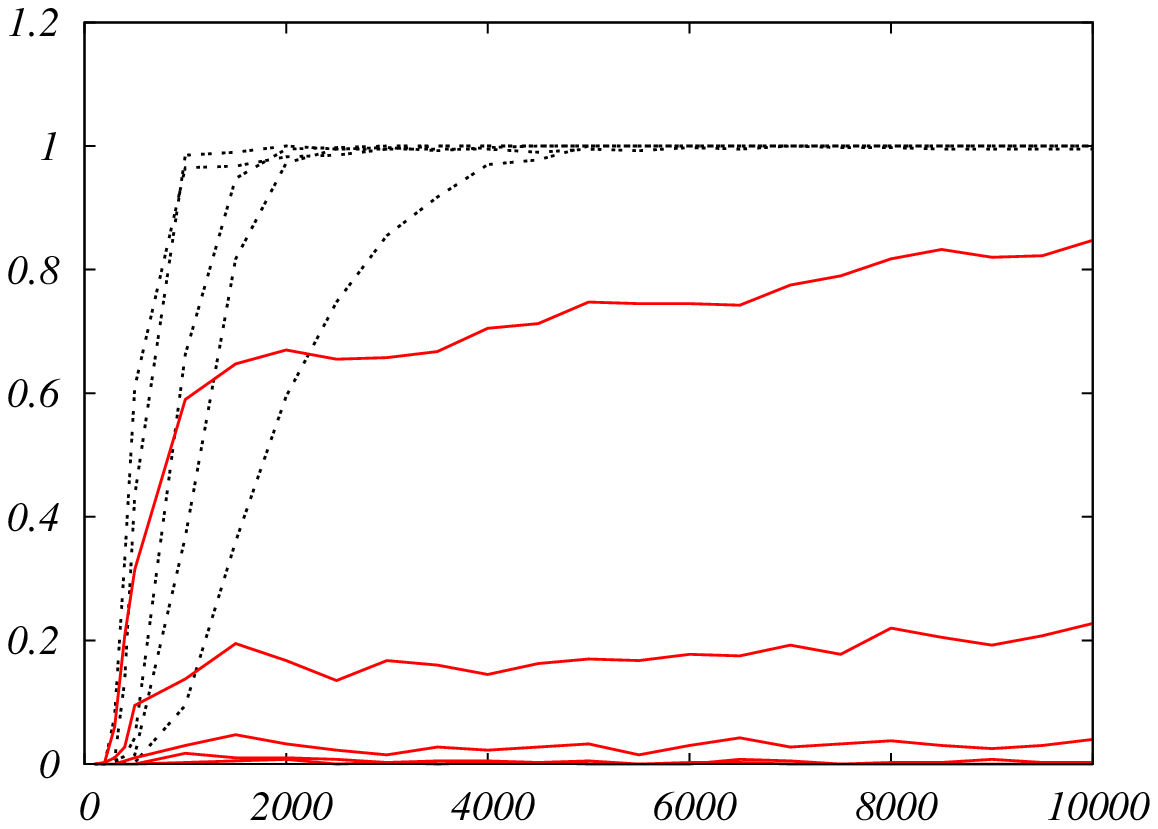}
\includegraphics[width=0.52\linewidth]{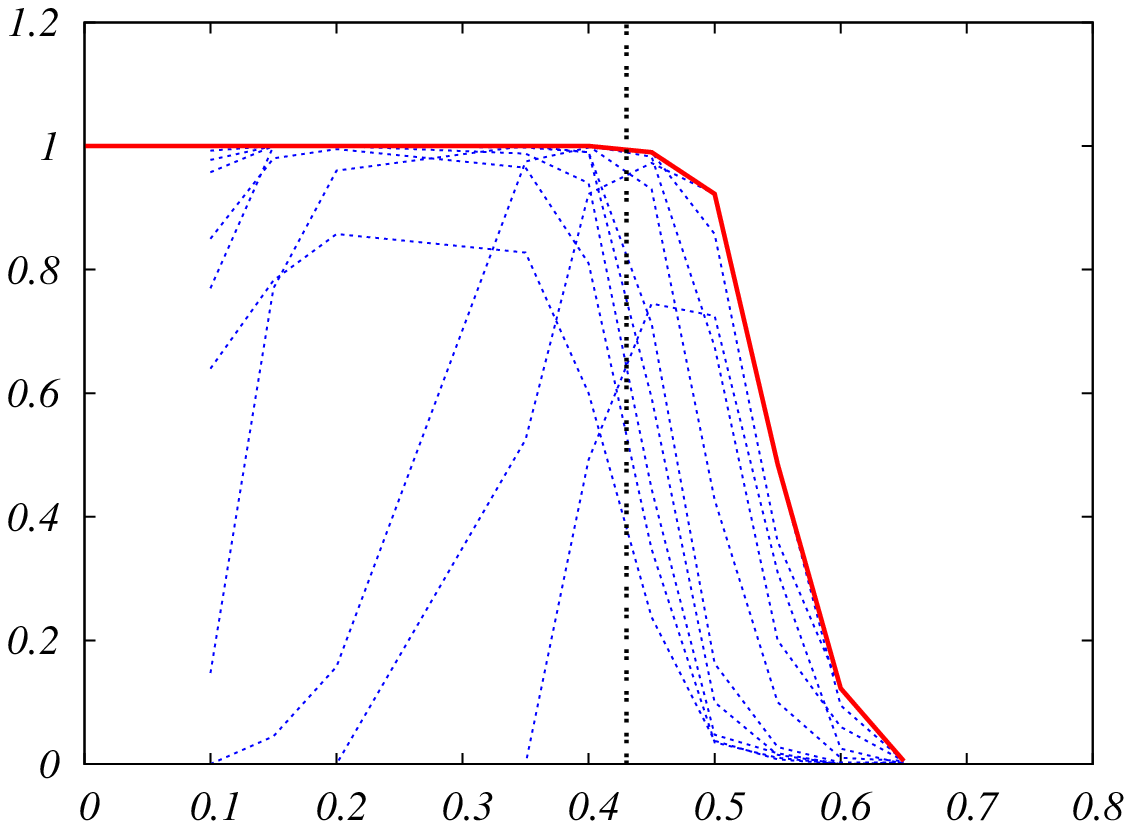}
\put(-410,65){P${}_{\rm succ}$}
\put(-380,118){\tiny{$\theta = 0.35,0.40$}}
\put(-367,107){\tiny{$\theta = 0.25$}}
\put(-363,100){\tiny{$\theta = 0.20$}}
\put(-350,90){\tiny{$\theta = 0.10$}}
\put(-300,93){\tiny{$\theta = 0.45$}}
\put(-300,35){\tiny{$\theta = 0.50$}}
\put(-350,18){\tiny{$\theta = 0.65,0.60,0.55$}}
\put(-310,-5){$n$}
\put(-100,-5){$\theta$}
\put(-92,17){$\theta_{thr}$}
\put(-200,65){P${}_{\rm succ}$}

\caption{Learning uniformly random graphs of degree 4
 from Ising models samples, using $\rlr$. Left: success probability as a
function of the number of samples $n$ for several values of $\theta$.
Right: the same data 
plotted for several choices of $\lambda$ versus $\theta$
as in Fig.~\ref{fig:Grid}, right panel.}\label{fig:RandomG}
\end{figure}

Figure \ref{fig:RandomG} presents similar data when $G$ 
is a uniformly random graph of degree  $\Delta=4$, over
$p=50$ vertices. The evolution of the success probability 
with $n$ clearly shows a dichotomy. When $\theta$ is below a 
threshold, a small number of samples is sufficient to reconstruct $G$ 
with high probability. Above the threshold even $n=10^4$ samples
are to few. In this case we can predict the threshold analytically,
cf. Lemma \ref{th:mart23} below, and get 
$\theta_{\rm thr}(\Delta=4)\approx 0.4203$, which compares favorably 
with the data.
%
%
\section{Proofs}

In order to prove Theorem \ref{th:mart2}, we need a few auxiliary results. 
It is convenient to introduce some notations. If $M$ is a matrix and 
$R,P$ are index sets then $M_{R\;P}$ denotes the submatrix 
with  row indices in $R$ and column indices in $P$.
As above, we let $r$ be the vertex whose neighborhood we are trying to 
reconstruct and define $S=\dr$, $S^c = V \setminus{\dr \cup r}$.  
Since the cost function $L (\utheta; \{x^{(\ell)}\})+\lambda||\utheta||_1$
only depend on $\utheta$ through its components $\utheta_{r,\cdot}=
\{\theta_{rj}\}$,
we will hereafter neglect all the other parameters and write
  $\utheta$ as a shorthand of $\utheta_{r,\cdot}$.

Let $\hz^*$  be a subgradient of $||\utheta||_1$  evaluated at the 
true parameters values, $\utheta^* = \{\theta_{rj}: \theta_{ij} = 0, \; 
\forall j \notin \dr, \theta_{rj} = \theta, \; \forall j \in \dr\}$. 
Let $\hutheta^n$ be the parameter estimate returned
by $\rlr(\lambda)$ when the number of samples is $n$. 
Note that, since we assumed
$\utheta^*\ge 0$,  $\hz^*_S = \mathbbm{1}$. Define 
$Q^n(\utheta,;\{\ux^{(\ell)} \})$ to be the Hessian 
of $L (\utheta; \{x^{(\ell)}\})$ and 
$Q(\utheta) = \lim_{n \rightarrow \infty} Q^n(\utheta,;\{\ux^{(\ell)}\})$. 
By the law of large numbers $Q(\utheta)$ is the Hessian
of  $\E_{G,\utheta}\log  \prob_{G,\utheta} (X_r| X_{\backslash r})$ where 
$\E_{G,\utheta}$ is the expectation with respect to (\ref{eq:IsingModel2}) and 
$\uX$ is a random variable distributed according to (\ref{eq:IsingModel2}). 
We will denote the maximum and minimum eigenvalue of a symmetric matrix 
$M$ by $\sigma_{\rm max}(M)$ and $\sigma_{\rm min}(M)$ respectively.

We will omit arguments whenever clear from the context. 
Any quantity evaluated at the true parameter values will be represented 
with a $^*$, e.g. $Q^* = Q(\theta^*)$. Quantities under a 
$\wedge$ depend on $n$.
Throughout this section $G$ is a graph of maximum degree $\Delta$.
%
%
\subsection{Proof of Theorem \ref{th:mart2}}\label{sec:ProofTheorem}

Our first auxiliary results establishes that, 
if $\lambda$ is small, then 
$||Q^*_{S^c S} {Q^*_{SS}}^{-1} \hz^*_S||_{\infty}>1$ is a sufficient condition
for the failure of $\rlr(\lambda)$.
\begin{lemma}
Assume $[ Q^*_{S^c S} {Q^*_{SS}}^{-1} \hz^*_S]_i \geq 1 + \epsilon$ for some 
 $\epsilon>0$ and some row $i\in V$, 
 $\sigma_{\rm min}(Q^*_{SS}) \ge  C_{\rm min} > 0$, and  
$\lambda < \sqrt{C_{\rm min}^3\eps /2^9 \Delta^4}$.
Then the success probability of $\rlr(\lambda)$ is upper bounded 
as
\begin{equation}
\psucc\le 4 \Delta^2  e^{-n \delta_A^2} +
 2\Delta\, e^{-n \lambda^2 \delta_B^2}
\label{th:mart21prob}
\end{equation}
where $\delta_A = (C_{\rm min}^2/100 \Delta^2) \epsilon$ and 
$\delta_B = (C_{\rm min} /8 \Delta) \epsilon$.
\label{th:mart21}
\end{lemma}
The next Lemma implies that, for $\lambda$ to be `reasonable'
(in the sense introduced in Section \ref{sec:Pseudo}),
$n\lambda^2$ must be unbounded.
\begin{lemma}
There exist $M = M(K,\theta)>0$ for $\theta>0$ such that the following is true: 
If $G$ is the graph with only one edge between nodes $r$ and
$i$ and  $n \lambda^2 \leq K$, then 
\begin{align}
\psucc\le  e^{-M(K,\theta)p}+ e^{-n(1-\tanh\theta)^2/32}\, .
\end{align}
\label{th:mart22}
\end{lemma}
\vspace{-0.1cm}

Finally, our key result shows that the condition 
$||Q^*_{S^c S} {Q^*_{SS}}^{-1} \hz^*_S||_{\infty}\le 1$ is violated
with high probability for large random graphs.
The proof of this result relies on a local weak convergence result
for ferromagnetic Ising models on random graphs proved in \cite{andrea}.
\begin{lemma}\label{th:mart23}
Let $G$ be a uniformly random regular graph of degree $\Delta>3$,
and $\epsilon>0$ be sufficiently small.
Then, there exists $\theta_{\rm thr}(\Delta,\eps)$
such that, for $\theta>\theta_{\rm thr}(\Delta,\eps)$,
 $|| Q^*_{S^c S} {Q^*_{SS}}^{-1}\hz^*_S||_{\infty} \ge 1 + \epsilon$ with 
probability converging to $1$ as  $p\to\infty$. 

Furthermore, for large $\Delta$, 
$\theta_{\rm thr}(\Delta,0+) = \tilde{\theta}\, \Delta^{-1}(1+o(1))$.
The constant $\tilde{\theta}$ is given by $\tilde{\theta} =
\tanh \bar{h})/\bar{h}$ and $\bar{h}$ is the unique positive solution of
$\bar{h} \tanh \bar{h} = (1-  \tanh^2 \bar{h})^2$.
Finally, there exist $C_{\rm min}>0$ dependent only on $\Delta$
and $\theta$ such that $\sigma_{\rm min}(Q^*_{SS})\ge C_{\rm min}$
 with 
probability converging to $1$ as  $p\to\infty$. 
\end{lemma}
The proofs of Lemmas \ref{th:mart21} and \ref{th:mart23} are 
sketched in the next subsection. Lemma \ref{th:mart22} is more 
straightforward and we omit its proof for space reasons.

\begin{proof}(Theorem \ref{th:mart2})
Fix $\Delta>3$, $\theta > K / \Delta$ (where $K$ is a large enough 
constant independent of $\Delta$), and $\eps,C_{\rm min}>0$ and both small enough. 
By Lemma \ref{th:mart23},
for any $p$ large enough we can choose a $\Delta$-regular
graph $G_p=(V=[p],E_p)$ and a vertex $r\in V$ such that 
$|Q^*_{S^c S} {Q^*_{SS}}^{-1} \mathbbm{1}_S|_i > 1 + \epsilon$ for 
some $i\in V\setminus r$.

By Theorem 1 in \cite{mossel} we can assume, without loss of generality
$n>K'\Delta\log p$ for some small constant $K'$. Further by
Lemma \ref{th:mart22}, $n\lambda^2\ge F(p)$ for some $F(p)\uparrow\infty$
as $p\to\infty$ and the condition of Lemma \ref{th:mart21} on $\lambda$ is satisfied since by the "reasonable" assumption $\lambda \to 0$ with $n$. Using these results in Eq.~(\ref{th:mart21prob}) 
of Lemma \ref{th:mart21} we get the following upper bound on the success probability
\begin{equation}
\psucc(G_p)\le 4 \Delta^2  p^{-\delta_A^2K'\Delta} +
 2\Delta\, e^{-n F(p) \delta_B^2}\, .
\end{equation}
In particular $\psucc(G_p)\to 0$ as $p\to\infty$.
\end{proof}

%
%
\subsection{Proofs of auxiliary lemmas}

\begin{proof}(Lemma \ref{th:mart21})
We will show that under the assumptions of the lemma and if 
$\hat{\utheta} = (\hat{\utheta}_S, \hat{\utheta}_{S^C}) = (\hutheta_S,0)$ 
then the probability that the $i$ component of any subgradient of 
$L (\utheta;\{x^{(\ell)}\}) + \lambda || \utheta ||_1$ vanishes
for any $\hutheta_S>0$ (component wise)
is upper bounded as in Eq.~\eqref{th:mart21prob}. To simplify notation 
we will omit $\{\ux^{(\ell)}\}$ in all the expression derived from $L$.

Let $\hz$ be a subgradient of $||\utheta||$ at $\hutheta$
and assume $\nabla L(\hutheta) + \lambda \hat{z}=0$. 
An application of the mean value theorem yields
\begin{equation}
\nabla^2 L (\utheta^*) [\hat{\utheta} - \utheta^*] = W^n  - \lambda \hat{z} + R^n\, ,
\end{equation}
where $W^n = -\nabla L (\utheta^*)$ and $[R^n]_j = [\nabla^2 L (\bar{\utheta}^{(j)}) - \nabla^2 L (\utheta^*)]_j^T (\hat{\utheta} - \utheta^*)$  with $\bar{\utheta}^{(j)}$ a point in the line from $\hat{\utheta}$ to $\utheta^*$. 
Notice that by definition $\nabla^2 L (\utheta^*) = {Q^n}^* = 
Q^n (\utheta^*) $. To simplify notation we will omit the $*$ in all ${Q^n}^*$. All $Q^n$ in this proof are thus evaluated at $\utheta^*$.

Breaking this expression into its $S$ and $S^c$ components and since 
$\hutheta_{S^C} =\utheta^*_{S^C} = 0$ we can eliminate $\hutheta_S - \utheta_S^*$ from the two expressions obtained and write
\begin{equation}
[W^n_{S^C} - R^n_{S^C}] - Q^n_{S^C S}(Q^n_{SS})^{-1} [W^n_S - R^n_S] + \lambda  Q^n_{S^C S}(Q^n_{SS})^{-1} \hat{z}_S = \lambda \hat{z}_{S^C}\, .
\end{equation}

Now notice that $Q^n_{S^C S}(Q^n_{SS})^{-1} = T_1 + T_2 + T_3 + T_4$ where
\begin{align*}
T_1 & =  Q^*_{S^CS}[(Q^n_{SS})^{-1} - (Q^*_{SS})^{-1}]\, ,
\;\;\;\;\;\;\;\;\;\;\;\;\;\;\;\;\;\;\;\;\;\;\;\;\;
T_2  =  [Q^n_{S^C S} - Q^*_{S^C S}] {Q^*_{SS}}^{-1}\, , \\
T_3 & =  [Q^n_{S^C S} - Q^*_{S^C S}] [(Q^n_{SS})^{-1} - (Q^*_{SS})^{-1}]\, ,
\;\;\;\;\;\;\;\;\;\;
T_4  =  Q^*_{S^CS} {Q^*_{SS}}^{-1}  \, .
\end{align*}
We will assume that the samples $\{x^{(\ell)}\}$ are such that 
the following event holds
\begin{equation}
\cE \equiv \{ ||Q^n_{SS} - Q^*_{SS}||_\infty < \xi_A , ||Q^n_{S^C S} - Q^*_{S^C S}||_\infty < \xi_B, ||W^n_S /\lambda||_\infty < \xi_C \}\, ,
\end{equation}
where $\xi_A \equiv C_{\rm min}^2\eps/(16 \Delta)$, 
$\xi_B \equiv C_{\rm min}\eps/(8 \sqrt{\Delta})$ and $\xi_C 
\equiv C_{\rm min}\eps/(8 \Delta)$.
Since $\mathbb{E}_{G, \theta} (Q^n) = Q^*$ and $\E_{G, \theta} (W^n) = 0$ and noticing that both $Q^n$ and $W^n$ are  sums of bounded i.i.d. random variables, a simple application of Azuma-Hoeffding inequality upper bounds
 the probability of 
$\cE$ as in \eqref{th:mart21prob}.

From $\cE$ it follows that $\sigma_{\rm min}(Q^n_{SS}) > \sigma_{\rm min}(Q^*_{SS})
 - C_{\rm min}/2 > C_{\rm min}/2$. We can therefore lower bound the absolute 
value of the $i^{\rm th}$ component of $\hat{z}_{S^C}$ by
\begin{equation}
|[Q^*_{S^C S} {Q^*_{SS}}^{-1}{\mathbbm{1}}_S]_i| - ||T_{1,i} ||_\infty - 
||T_{2,i} ||_\infty - ||T_{3,i} ||_\infty - 
\Big| \frac{W^n_i}{\lambda} \Big| - \Big| \frac{R^n_i}{\lambda} 
\Big| - \frac{\Delta}{C_{\rm min}} \left( \Big|\Big| 
\frac{W^n_S}{\lambda} \Big|\Big|_\infty + \Big|\Big| \frac{R^n_S}{\lambda} 
\Big|\Big|_\infty  \right) \, ,\nonumber
\end{equation}
where the subscript $i$ denotes the $i$-th row of a matrix.

The proof is completed by showing that the event 
$\cE$ and the assumptions of the theorem 
imply that each of last $7$ terms in this expression is smaller than 
$\epsilon/8$. Since $|[Q^*_{S^C S} {Q^*_{SS}}^{-1}]_i^T \hat{z}^n_S|\ge
1+\eps$ by assumption, this implies  $|\hat{z}_i|\ge 1 + \epsilon/8 > 1$ 
which cannot be since any subgradient of the $1$-norm has components of 
magnitude at most $1$. 

The last condition on $\cE$ immediately bounds all terms involving $W$ by 
$\epsilon/8$. Some straightforward manipulations imply (See Lemma 7 from \cite{martin})
\begin{align*}
||T_{1,i} ||_\infty & \leq \frac{\Delta}{C_{\rm min}^2} ||Q^n_{SS} - Q^*_{SS}||_\infty \, ,\;\;\;\;\;\;\;\;\;
||T_{2,i} ||_\infty  \leq \frac{\sqrt{\Delta}}{C_{\rm min}} ||[Q^n_{S^C S} - Q^*_{S^C S}]_i||_\infty\, , \\
||T_{3,i} ||_\infty & \leq \frac{2 \Delta}{C_{\rm min}^2} ||Q^n_{SS} - Q^*_{SS}||_\infty   ||[Q^n_{S^C S} - Q^*_{S^C S}]_i||_\infty \, ,
\end{align*}
and thus all will be bounded by $\epsilon/8$ when $\cE$ holds.
The upper bound of $R^n$ follows along similar lines via an 
mean value theorem, and is deferred to a longer version of this paper.
\end{proof}

\begin{proof}(Lemma \ref{th:mart23}.)
Let us state explicitly the local weak convergence result mentioned in
Sec.~\ref{sec:ProofTheorem}. For $t\in\naturals$, let $\Tree(t)=
(V_\Tree,E_{\Tree})$ be the regular rooted tree of $t$ generations and 
define the associated Ising measure as
\vspace{-0.1cm}
\begin{align}
\mu_{\Tree,\theta}^+(\ux) =\frac{1}{Z_{\Tree,\theta}}\,
\prod_{(i,j)\in E_{\Tree}} e^{\theta x_i x_j}\prod_{i\in \dtree(t)}e^{h^*x_i}\, .
\label{IsingTree}
\end{align}
Here $\dtree(t)$ is the set of leaves of $\Tree(t)$ and
 $h^*$ is the unique positive solution of
$h = (\Delta -1)\, \atanh\, \{\tanh \theta \, \tanh  h \}$.
It can be proved using \cite{andrea} and uniform continuity 
with respect to the `external field' that non-trivial local expectations
with respect to $\mu_{G,\theta}(\ux)$ converge to local expectations
with respect to $\mu_{\Tree,\theta}^+(\ux)$, as $p\to\infty$.

More precisely, let 
$\Ball_r(t)$ denote a ball of radius $t$ around node $r\in G$
(the node whose neighborhood we are trying to reconstruct).
For any fixed $t$, the probability that $\Ball_r(t)$ is not
isomorphic to $\Tree(t)$ goes to $0$ as $p\to\infty$. 
Let $g(\ux_{\Ball_r(t)})$ be any function of the variables in
$\Ball_r(t)$ such that $g(\ux_{\Ball_r(t)}) = g(-\ux_{\Ball_r(t)})$.
Then  almost surely over graph sequences $G_p$ of uniformly
random regular graphs with $p$ nodes (expectations here are 
taken with respect to the measures (\ref{eq:IsingModel}) and 
(\ref{IsingTree}))
\begin{eqnarray}
\lim_{p\to\infty}\E_{G,\theta} \{g(\uX_{\Ball_r(t)})\}
=\E_{\Tree(t),\theta,+} \{g(\uX_{\Tree(t)})\}\, .\label{eq:Weak}
\end{eqnarray}
The proof consists in considering 
$[ Q^*_{S^c S} {Q^*_{SS}}^{-1} \hat{z}^*_S]_i$ for $t=\dist(r,i)$ finite.
We then write  $(Q^*_{SS})_{lk} = \E\{ g_{l,k}(\uX_{_{\Ball_r(t)}})\}$
and $(Q^*_{S^cS})_{il} = \E\{ g_{i,l}(\uX_{_{\Ball_r(t)}})\}$ for some functions
$g_{\cdot,\cdot}(\uX_{_{\Ball_r(t)}})$ and apply the weak convergence result
(\ref{eq:Weak}) to these expectations. We thus reduced the calculation 
of  $[ Q^*_{S^c S} {Q^*_{SS}}^{-1} \hat{z}^*_S]_i$ to the calculation 
of expectations with respect to the tree measure (\ref{IsingTree}).
The latter can be implemented explicitly through a recursive procedure,
with simplifications arising thanks to the tree symmetry and by
taking $t\gg 1$. The actual calculations consist in a (very) long
exercise in calculus and we omit them from this outline.

The lower bound on $\sigma_{\rm min}(Q^*_{SS})$ is proved by a similar 
calculation.
\end{proof}
\vspace{-0.35cm}

\subsubsection*{Acknowledgments}

This work was partially supported by a Terman fellowship, the NSF CAREER award CCF-0743978 and the NSF grant DMS-0806211 and by a Portuguese Doctoral FCT fellowship.

%

\newpage

\bibliographystyle{amsalpha}

\begin{thebibliography}{99}


\bibitem{abbeel} P. Abbeel, D. Koller and A. Ng, ``Learning factor graphs in polynomial time and sample
complexity''. Journal of Machine Learning Research., 2006, Vol. 7, 1743--1788.

\bibitem{martin_info_limits} M. Wainwright, ``Information-theoretic limits on sparsity recovery in the high-dimensional and noisy setting'', arXiv:math/0702301v2 [math.ST], 2007.

\bibitem{santhanam_info_limits} N. Santhanam, M. Wainwright, ``Information-theoretic limits of selecting binary graphical models in high dimensions'', 	arXiv:0905.2639v1 [cs.IT], 2009.

\bibitem{mossel} G. Bresler, E. Mossel and A. Sly, ``Reconstruction of Markov Random Fields from Samples:
Some Observations and Algorithms'',Proceedings of the 11th international workshop,
APPROX 2008, and 12th international workshop RANDOM 2008, 2008 ,343--356.

\bibitem{Csiszar} Csisz�ar and Z. Talata, ``Consistent estimation of the basic neighborhood structure of Markov random fields'', The Annals of Statistics, 2006, 34, Vol. 1, 123-�145.

\bibitem{Friedman} N. Friedman, I. Nachman, and D. Peer, ``Learning
Bayesian network structure from massive datasets: The
�sparse candidate� algorithm''. In UAI, 1999.

\bibitem{martin} P. Ravikumar, M. Wainwright and J. Lafferty, ``High-Dimensional Ising Model Selection Using
l1-Regularized Logistic Regression'', arXiv:0804.4202v1  [math.ST], 2008.

\bibitem{martin2} M.Wainwright, P. Ravikumar, and J. Lafferty, ``Inferring
graphical model structure using l1-regularized pseudolikelihood``, In NIPS, 2006.

\bibitem{Tibshirani} H. H\"{o}fling and R. Tibshirani, 
``Estimation of Sparse Binary Pairwise Markov
Networks using Pseudo-likelihoods'' , Journal of Machine Learning Research, 2009, Vol. 10, 883--906.

\bibitem{Ghaoui} O.Banerjee, L. El Ghaoui and A. d'Aspremont, ``Model Selection Through Sparse Maximum Likelihood Estimation for Multivariate Gaussian or Binary Data'', Journal of Machine Learning Research, March 2008, Vol. 9, 485--516.

\bibitem{Yuan} M. Yuan and Y. Lin, ``Model Selection and Estimation
in Regression with Grouped Variables'', J. Royal. Statist. Soc B, 2006, 68, Vol. 19,49--67.

\bibitem{Meinshausen} N. Meinshausen and P. B\"{u}uhlmann, ``High dimensional graphs and variable selection with
the lasso'', Annals of Statistics, 2006, 34, Vol. 3.

\bibitem{tibshirani_lasso} R. Tibshirani, ``Regression shrinkage and selection via the lasso'', Journal of the Royal Statistical Society, Series B, 1994, Vol. 58, 267--288.

\bibitem{zhao} P. Zhao, B.  Yu, ``On model selection consistency of Lasso'', Journal of Machine. Learning Research 7, 2541–2563, 2006.

\bibitem{Zobin} D. Zobin, ''Critical behavior of the bond-dilute two-dimensional 
Ising model``, Phys. Rev., 1978 ,5, Vol. 18, 2387 -- 2390.

\bibitem{fisher} M. Fisher, ''Critical Temperatures of Anisotropic Ising Lattices. II. General Upper
Bounds'', Phys. Rev. 162 ,Oct. 1967, Vol. 2, 480--485.

\bibitem{andrea} A. Dembo and A. Montanari, ``Ising Models on Locally Tree Like Graphs'', Ann. Appl. Prob. (2008), to appear, arXiv:0804.4726v2 [math.PR]


\end{thebibliography}

\end{document}